\documentclass[twoside,twocolumn,a4paper,11pt]{article}
\usepackage{colacl}
\usepackage{latexsym}
\newlength{\lwidth}
\newlength{\dst}
\newlength{\src}
\newlength{\innerlen}
\newlength{\arrowlen}
\newcommand{\link}[5]{
\settowidth{\dst}{#2}%
\settowidth{\innerlen}{#3}%
\settowidth{\src}{#4}%
\setlength{\arrowlen}{.5\dst}%
\addtolength{\arrowlen}{.5\src}%
\addtolength{\arrowlen}{\innerlen}%
#1%
\makebox[0pt][l]{\hspace{.5\dst}\rule[-2ex]{\arrowlen}{\lwidth}}%
\makebox[0pt][l]{\hspace{.5\dst}\rule[-2ex]{\lwidth}{1ex}}%
\makebox[0pt][l]{\rule[-1ex]{\dst}{\lwidth}}#2%
#3%
\makebox[0pt][l]{\hspace{.5\src}\rule[-2ex]{\lwidth}{1ex}}%
\makebox[0pt][l]{\rule[-1ex]{\src}{\lwidth}}#4%
#5%
}%

\begin{document}
\title{\textbf{Bootstrapping Structure using Similarity}}
\author{Menno~van~Zaanen\\
University of Leeds
}
\date{}
\maketitle

\begin{abstract}
In this paper a new similarity-based learning algorithm, inspired by string
edit-distance \cite{bib:tstscp}, is applied to the problem of bootstrapping
structure from scratch. The algorithm takes a corpus of unannotated sentences
as input and returns a corpus of bracketed sentences. The method works on
pairs of unstructured sentences or sentences partially bracketed by the
algorithm that have one or more words in common. It finds parts of sentences
that are interchangeable (i.e. the parts of the sentences that are different
in both sentences). These parts are taken as possible constituents of the same
type. While this corresponds to the basic bootstrapping step of the algorithm,
further structure may be learned from comparison with other (similar) sentences.

We used this method for bootstrapping structure from the flat sentences of the
Penn Treebank ATIS corpus, and compared the resulting structured sentences to
the structured sentences in the ATIS corpus. Similarly, the algorithm was
tested on the OVIS corpus. We obtained 86.04~\% non-crossing brackets
precision on the ATIS corpus and 89.39~\% non-crossing brackets precision on
the OVIS corpus.
\end{abstract}

\section{Introduction}

People seem to learn syntactic structure without great difficulty.
Unfortunately, it is difficult to model this process and therefore, it is
difficult to make a computer learn structure.

Instead of actually modeling the human process of language learning, we propose
a grammar learning algorithm and apply it to a corpus of natural language
sentences. The algorithm should assign structure to the sentences which is
similar to the structure people give to the sentences.

The algorithm consists of two phases. The first phase, \emph{alignment
learning}, builds a large set of possible constituents by aligning sentences.
The second phase, which is called \emph{bracket selection}, selects the best
constituents from this set.

The rest of the paper is organized as follows. We will start out by describing
some previous work in grammar induction. This is followed by a detailed
description of the ABL algorithm, where three different implementations of ABL
will be described, \emph{ABL:incr}, \emph{ABL:leaf} and \emph{ABL:branch}.
We will then discuss some of the properties of the ABL algorithms. First, we
claim the ABL algorithms can generate recursive structures and then discuss
problems with PP attachment. After that, some results of the algorithms applied to the ATIS corpus \cite{bib:balacoetpt} and to the OVIS corpus
\cite{bib:admfsi} will be presented, followed by a description of future
research on the ABL algorithm.

\section{Previous Work}

This section contains a brief overview of previous work in grammar acquisition
systems. Some differences between supervised and unsupervised acquisition
methods will be discussed, followed by a discussion of several different
methods. We then relate ABL to some of these methods.

\subsection{Supervised versus Unsupervised}

Language learning algorithms can be roughly divided into two groups,
\emph{supervised} and \emph{unsupervised} learning algorithms based on the
amount of information about the language they use. All learning algorithms use
a \emph{teacher} that gives examples of (unstructured) sentences in the
language.  In addition, some algorithms also use a \emph{critic}. A critic may
be asked if a certain sentence (possibly including structure) is a valid
sentence in the language. The algorithm can use a critic to validate
hypotheses about the language.\footnote{When an algorithm uses a treebank or
structured corpus to initialise, it is said to be supervised. The
structure of the sentences in the corpus can be seen as the critic.} Supervised
language learning methods use a teacher and a critic, whereas the unsupervised
methods only use a teacher. \cite{bib:mlofnl}

Supervised language learning methods typically generate better results. These
methods can tune their output, since they receive knowledge of the structure
of the language (by initialisation or querying a critic). In contrast,
unsupervised language learning methods do not receive these structured
sentences, so they do not know at all what the output should look like.

However, it is worthwhile to investigate unsupervised language learning methods,
since ``the costs of annotation are prohibitively time and expertise
intensive, and the resulting corpora may be too susceptible to restriction to a
particular domain, application, or genre''. \cite{bib:ulinlp}

\subsection{Language Learning Methods}

In this section a short overview of several supervised and unsupervised
language learning methods will be given. However, the focus will be on on
unsupervised language learning methods, since we are interested in what
results can be achieved using as little information as possible.

There have been many different approaches to language learning, only some of
the different systems will be described here briefly. This section is merely
meant to relate the ABL system to other language learning methods; we do
not claim to give a complete overview of this field.

A lot of recent language learning methods are based on probabilistic or
counting theories. An example of this type of method is Memory-Based Learning
(MBL) which keeps track of the distribution of contexts of words and assigns
word types based on that information \cite{bib:mblaap}. \newcite{bib:pnlumis}
describe a system that can find constituent boundaries using mutual
information of n-grams within sentences, while in \cite{bib:bscusm} and
\cite{bib:diapcfasc} models are proposed that use distributional information to
acquire syntactic categories.

Another type of language learning system is based on the Minimum Description
Length (MDL) principle. These systems compress the input corpus by finding the
minimum description needed to express the corpus. The compression results in a
grammar that can describe the corpus. Examples of these systems can be found in
\cite{bib:giatmdlp} and \cite{bib:ula}.

\newcite{bib:ladcag} describes a system that performs a heuristic search
while creating and merging symbols directed by an evaluation function.
Similarly, \newcite{bib:gibhc} describe an algorithm that uses a cost
function to direct search for a grammar. The grammar induction method found in
\cite{bib:ipgbbmm} merges elements of models using a Bayesian framework.
\newcite{bib:bgiflm} presents a Bayesian grammar induction method, which is
followed by a post-pass using the inside-outside algorithm
\cite{bib:tgfsr,bib:teoscfgutioa}, while \newcite{bib:iorfpbc} apply the
inside-outside algorithm to a partially structured corpus.

The final system mentioned here is Transformation-Based Learning
\cite{bib:agiapftatba}, which is also a supervised system. This system differs
from others in that it is a non-probabilistic system that learns
transformations to improve a naive parse (for example a right branching parse).

\subsection{ABL in relation to other Methods}

ABL consists of two distinct phases, alignment learning and bracket selection.
Both phases can be roughly compared to different language learning methods.

The alignment learning step is a way of compressing the corpus. Similar
techniques can be found in systems that are based on the MDL principle
\cite{bib:giatmdlp}. MDL systems compress the corpus by looking for the
minimal description of the corpus. The MDL principle results in grouping
reoccurring parts of sentences yielding a reduction of the corpus. The
alignment learning step finds constituents based on the idea of
interchangeability, which effectively compresses the corpus.

The bracket selection phase selects constituents based on the probability of the
possible constituents. This is in some way similar to systems that use
distributional information to select the most probable syntactic types as in
\cite{bib:bscusm} or \cite{bib:diapcfasc}. On the other hand, ABL assigns a
probability to the different constituents, which is similar to an SCFG
\cite{bib:tgfsr}.

\section{Algorithm}

In this section we describe an unsupervised grammar learning algorithm that
learns using a corpus of plain sentences, so neither pre-tagged nor pre-labelled
or bracketed sentences are used. The output of the algorithm is a labelled,
bracketed version of the input corpus. Although the algorithm does not generate
a (context-free) grammar, it is trivial to deduce one from the output treebank.

The algorithm consists of two distinct phases: \emph{alignment learning} and
\emph{bracket selection}. Alignment learning finds possible constituents by
aligning sentences from the corpus. The bracket selection phase selects 
constituents from the possibly overlapping constituents that were found in the
alignment learning phase.

Both phases will be described in more detail in the next two sections.

\subsection{Alignment Learning}

The alignment learning phase is based on Harris's idea of interchangeability
\cite{bib:misl} that states that \emph{constituents of the same type can be
replaced by each other}. This means that for example in the sentence
\textit{What is a dual carrier} the noun phrase \textit{a dual carrier} may be
replaced by another noun phrase.\footnote{Most example sentences used in this
paper can be found in the ATIS corpus.} Replacing the noun phrase with
\textit{the payload of an African Swallow} yields the sentences \textit{What is
the payload of an African Swallow}, which is another well-formed sentence. 

ABL uses the replacement feature to find constituents by reversing the idea.
Instead of replacing parts of sentences that have the same type, this idea is
used to find constituents of the same type by looking for parts of sentences
that can be replaced by each other. In the example this means the algorithm
looks for parts as \textit{a dual carrier} and \textit{the payload of an
African Swallow}, which then are remembered as possible constituents.

Finding replaceable parts of sentences is done by finding parts that are
identical in two sentences and parts that are distinct. The distinct parts
of the sentences are parts that can be interchanged. See for example the
sentences in figure~\ref{fig:bootstrapping}. The part \textit{What is} is
identical in both sentences, while the noun phrases are distinct parts. ABL now
assumes the distinct parts of the sentences are constituents of the same type.

\begin{figure}[ht]
\begin{center}
\begin{tabular}{l}
\textit{What is a dual carrier}\\
\textit{What is the payload of an African Swallow}\\[.05cm]
\hline
\textit{What is (a dual carrier)$_{X_1}$}\\
\textit{What is (the payload of an African Swallow)$_{X_1}$}\\[.05cm]
\end{tabular}
\end{center}
\caption{Bootstrapping structure}
\label{fig:bootstrapping}
\end{figure}

An instance of the string edit distance algorithm \cite{bib:tstscp} that finds
the longest common subsequences in two sentences is used. These longest common
subsequences are parts that are identical in both sentences. The
\emph{distinct} parts of the sentences are exactly these parts of the
sentences that are not part of the longest common subsequences. In
figure~\ref{fig:bootstrapping}, \textit{What is} is the longest common
subsequence (and the only one), while \textit{a dual carrier} and \textit{the
payload of an African Swallow} are the remaining (distinct) parts of the
sentences.

The distinct parts of the sentences, found by the string edit distance
algorithm, are then grouped and labelled. Respective distinct parts receive
the same non-terminal. Since the algorithm does not know any linguistically
motivated names for the non-terminals, it assigns names $X_1, X_2, \ldots$ to
the different constituents.

\begin{figure}
\begin{tabbing}
For\=For\=For\=\kill
For each sentence $s_1$ in the corpus:\\
\>For every other sentence $s_2$ in the corpus:\\
\>\>Align $s_1$ to $s_2$\\
\>\>Find the identical and distinct parts\\
\>\>\>between $s_1$ and $s_2$\\
\>\>Assign non-terminals to the constituents\\
\>\>\>(i.e. distinct parts of $s_1$ and $s_2$)
\end{tabbing}
\caption{Alignment learning algorithm}
\label{fig:algorithm}
\end{figure}

More structure is learned when aligning more than two sentences. Each new
sentence is compared to all (partially structured) sentences in the partially
structured corpus. An overview of the algorithm can be found in
figure~\ref{fig:algorithm}.

Sentences are always aligned without looking at the structure that is already
known; all newly learned structure is added to the old structure, which may
yield overlapping constituents.

\begin{figure}[ht]
\begin{center}
\begin{tabular}{l}
\textit{(Book Delta 128)$_{X_1}$ from Dallas to Boston}\\
\textit{(Give me all flights)$_{X_1}$ from Dallas to Boston}\\\hline
\textit{Give me (all flights from Dallas to Boston)$_{X_2}$}\\
\textit{Give me (information on reservations)$_{X_2}$}
\end{tabular}
\end{center}
\vspace{-.3cm}
\caption{Overlapping constituents}
\label{fig:overlap}
\end{figure}

In figure~\ref{fig:overlap} two overlapping constituents are learned on the
sentence \textit{Give me all flights from Dallas to Boston}. Constituent $X_1$
is learned when aligning to the sentence \textit{Book Delta 128
from Dallas to Boston} and the other constituent ($X_2$) is learned when
aligning to the sentence \textit{Give me information on reservations}. The
bracket selection phase selects brackets until no more overlapping brackets are
found.

An unstructured sentence is sometimes aligned to a partially structured
sentence (which was already in the partially structured corpus). Aligning these
sentences might yield a constituent that was already present in the partially
structured sentence. The new constituent in the unstructured sentence then
receives the same non-terminal as the constituent in the partially structured
sentence. An example of this can be found in figure~\ref{fig:learn1}.
Sentences~1 and 2 are compared, resulting in sentences~3 and 4.

\begin{figure}[ht]
\begin{center}
\begin{tabular}{l}
\textbf{1} \textit{What does (AP57 restriction)$_{X_1}$ mean}\\
\textbf{2} \textit{What does aircraft code D8S mean}\\\hline
\textbf{3} \textit{What does (AP57 restriction)$_{X_1}$ mean}\\
\textbf{4} \textit{What does (aircraft code D8S)$_{X_1}$ mean}\\[.5cm]
\end{tabular}
\end{center}
\vspace{-.3cm}
\caption{Learning with a partially structured sentence and an unstructured
sentence}
\label{fig:learn1}
\end{figure}

It may even be the case that two partially structured sentences are aligned.
This occurs when a new sentence has been aligned to a sentence (and has received
some structure) and is then aligned to another partially structured sentence in
memory. If this combination of sentences yields a constituent with two distinct
non-terminals (a different one in each sentence), the two non-terminals are
merged. All occurrences of these non-terminals are updated in the corpus. In
figure~\ref{fig:learn2} sentences~1 and 2 are compared, resulting in
sentences~3 and 4.

\begin{figure}[ht]
\begin{center}
\begin{tabular}{l}
\textbf{1} \textit{Explain the (meal code)$_{X_1}$}\\
\textbf{2} \textit{Explain the (restriction AP)$_{X_2}$}\\\hline
\textbf{3} \textit{Explain the (meal code)$_{X_3}$}\\
\textbf{4} \textit{Explain the (restriction AP)$_{X_3}$}\\
\end{tabular}
\end{center}
\vspace{-.2cm}
\caption{Learning with two partially structured sentences}
\label{fig:learn2}
\end{figure}

Merging non-terminals (as shown in figure~\ref{fig:learn1} reduce the number of
different non-terminals. By merging non-terminals, we assume that constituents
in a certain context can only have one type. In section~\ref{s:ac} we discuss
the consequences and propose a method that loosens this assumption.

\subsection{Bracket Selection}

The alignment learning phase may generate unwanted overlapping constituents as
can be seen in figure~\ref{fig:overlap}. Since we assume the underlying
grammar of the corpus is context-free and we want to know the most appropriate
disambiguated structure of the sentences of the corpus, overlapping
constituents have to be eliminated. The bracket selection phase does exactly
this.

Three different approaches to the selection of constituents have been
implemented. Note that the selected constituents are kept (i.e. they are
\emph{not} removed).
\begin{description}
\item [ABL:incr] Assume the constituent learned earlier is correct. This means
that when new constituents overlap with older ones, they are ignored.
\item [ABL:leaf] Constituents are selected based on their probability. The
system computes the probability of a constituent by counting the number of
times the sequence of words in the constituent occurs as a constituent,
normalized by the total number of constituents. The probability of a
constituent $c$ can be computed as follows:
\end{description}
\begin{eqnarray*}
\lefteqn{P_{leaf}(c)=\frac{|c'\in C:yield(c')=yield(c)|}{|C|}}\\
&&\mbox{where $C$ is the entire set of constituents.}
\end{eqnarray*}
\begin{description}
\item [ABL:branch] This method is similar to the ABL:leaf method. The
probability of a constituent is now computed by counting the number of times the
sequence of words in a constituent \emph{together with its root non-terminal}
occur, normalized by the number of constituents with that root non-terminal:
\end{description}
\begin{eqnarray*}
\lefteqn{P_{branch}(c|root(c)=r)=}\\
&&\frac{|c'\in C:yield(c')=yield(c)\wedge root(c')=r|}{|c''\in C:root(c'')=r|}\\
\end{eqnarray*}

The ABL:incr method may be applied during the alignment learning phase. When a
constituent is found that overlaps with an existing constituent, the new
constituent will not be stored. ABL:leaf and ABL:branch bracket selection
methods are applied after the alignment learning phase, since more specific
constituent counts are available then.

The probabilistic methods, ABL:leaf and ABL:branch, both compute the probability
of constituents. Since more than just two constituents can overlap, the methods
need to consider the probability of all possible combinations of constituents,
which is the product of the probabilities of the separate constituents as in
SCFGs (cf. \cite{bib:profl}). Viterbi style algorithm optimization
\cite{bib:ebfccaaaoda} is used to efficiently select the best combination of
constituents.

Computing the probability of a combination of constituents by taking the
product of the probabilities leads to a ``thrashing'' effect. Since the product
of two probabilities is always smaller than or equal to the two single
probabilities, the system has a preference to single constituents over a
combination of constituents. Therefore, the geometric mean is used to compute
the probability of a combination of constituents instead
\cite{bib:nfomfbfpcp}.\footnote{The geometric mean of a set of constituents
$c_1, \ldots, c_n$ is \(P(c_1\wedge\ldots\wedge c_n)=\sqrt[n]{\prod_{i=1}^n
P(c_i)}\)} When more combinations of constituents have the same
probability, one is chosen at random.

\begin{figure*}[ht]
\begin{center}
\begin{tabular}{l}
\makebox[1.35cm][l]{\textbf{learned}}
\textit{Book reservations for five from Dallas to Baltimore on (flight 314 on
(May 12th)$_{X_{15}}$)$_{X_{15}}$}\\
\makebox[1.35cm][l]{\textbf{original}}
\textit{Book reservations for five from Dallas to Baltimore on (flight
314)$_{NP}$ on (May 12th)$_{NP}$}\\
\makebox[1.35cm][l]{\textbf{learned}}
\textit{What is the (name of the (airport in Boston)$_{X_{18}}$)$_{X_{18}}$}\\
\makebox[1.35cm][l]{\textbf{original}}
\textit{What is (the name of (the airport in Boston)$_{NP}$)$_{NP}$}\\
\makebox[1.35cm][l]{\textbf{learned}}
\textit{Explain classes QW and (QX and (Y)$_{X_{52}}$)$_{X_{52}}$}\\
\makebox[1.35cm][l]{\textbf{original}}
\textit{Explain classes ((QW)$_{NP}$ and (QX)$_{NP}$ and (Y)$_{NP}$)$_{NP}$}\\
\end{tabular}
\end{center}
\vspace{-.2cm}
\caption{Recursion in the ATIS corpus}
\label{fig:recursion}
\end{figure*}

\section{Discussion}

\subsection{Recursion}

All ABL algorithms generate recursive structures. Figure~\ref{fig:recursion}
shows some real examples of recursive structures generated by the ABL
systems compared to their structure in the ATIS corpus.

In the first example ABL finds a recursive structure, but there is no recursion
in the original sentence. The context of the constituents is quite similar
(both follow the word ``on''). The second example contains recursive
structures that are very alike. The original corpus contains two recursive
noun phrases, while the learned structure describes recursive noun phrases
without the determiner. The last example contains recursive noun phrases with
conjunction. Again similar structures are found in the learned and original
corpora.

Intuitively, a recursive structure is formed first by building the constituents
that form the structure of the recursion but with different root non-terminals.
Now the non-terminals need to be merged. This happens when two partially
structured sentences are compared to each other yielding a constituent that
already existed in both sentences with the non-terminals present in the
``recursive'' structure (see figure~\ref{fig:learn2}). The non-terminals are
then merged, resulting in a recursive structure.

\subsection{PP attachment}

\begin{figure}
\begin{tabular}{l}
\textbf{1} \textit{\link{The man }{saw}{ the girl }{with his squinting
eyes}}\\[.3cm]
\textbf{2} \textit{\link{The man saw }{the girl}{ }{with the bikini}{}}\\[.3cm]
\textbf{3} \textit{The man saw the girl with the binoculars}
\end{tabular}
\caption{PP attachment}
\label{fig:attachment}
\end{figure}

Most unsupervised grammar learning systems have difficulty regarding
prepositional phrase attachments. In figure~\ref{fig:attachment} the first two
sentences seem similar, but the prepositional phrase \textit{with his squinting
eyes} indicates the man \textit{saw} the girl using his eyes, while
\textit{with the bikini} determines \textit{the girl} to have the
bikini, since it is difficult to see using a bikini.\footnote{The phrase
\textit{with his squinting eyes} could also mean that the girl has squinting
eyes, just like the man, but this reading is less likely.} On the other hand,
the third sentence in figure~\ref{fig:attachment} is ambiguous.

To solve the problem of PP attachment, several types of information may be
needed. The PP attachment in the first sentence can be solved using syntactic
information only, since \textit{his} clearly indicates the man is looking.
However, the second sentence can only be solved using semantic information.
Without knowing that a bikini does not help enhancing sight, it is difficult if
not impossible to attach the prepositional phrase to the correct constituent.

The third sentence shows that sometimes discourse information is needed to
solve attachment. The prepositional phrase \textit{with the binoculars} should
be attached to \textit{saw} or \textit{the girl} depending on the who has the
binoculars.

Since the ABL system builds structures based on a context-free grammar and
does not use any other information than the input sentences, ABL might seem 
unable to solve PP attachments correctly. However, the bracket selection phase
might find a difference in distribution between \textit{the girl with his
squinting eyes} and \textit{the girl with the bikini}. This difference in
distribution might resolve in the different PP attachments.

\section{Results}
\label{s:numerical}

This section describes the results of applying the three ABL algorithms
and two baseline systems to the Penn Treebank ATIS (Air Travel Information
System) corpus \cite{bib:balacoetpt} and to the OVIS corpus
\cite{bib:admfsi}.\footnote{OVIS (Openbaar Vervoer Informatie Systeem) stands
for Public Transport Information System.} First the test environment is
described, followed by a discussion of the test results.

\subsection{Test Environment}

The three ABL algorithms, ABL:incr, ABL:leaf and ABL:branch have been applied to
the ATIS and OVIS corpus. The ATIS corpus contains 716 sentences and 11,777
constituents. The larger OVIS corpus contains 10,000 sentences. When single
word sentences are removed, this results in a corpus of 6,797 sentences
containing 48,562 constituents.

The sentences of the corpora are stripped of their structure. The resulting
plain sentences are used in the learning algorithms and the resulting
structured sentences are then compared to the structure of the sentences in the
original corpora.

To be able to compare the results, we have also computed the results of two
baseline systems, a left-branching and a right-branching system on both
corpora.

\begin{figure*}[ht]
\[NCBP=\frac{\sum_i|NonLaBr(O_i)|-|Cross(NonLaBr(O_i), NonLaBr(T_i))|}
{\sum_i|NonLaBr(O_i)|}\]
\[NCBR=\frac{\sum_i|NonLaBr(T_i)|-|Cross(NonLaBr(T_i), NonLaBr(O_i))|}
{\sum_i|NonLaBr(T_i)|}\]
\[ZCS=\frac{\sum_i Cross(O_i, T_i)=0}{|TEST|}\]
\vspace{.3cm}
\begin{center}
\begin{tabular}{l}
$NonLaBr(T)$ denotes the set of non-labelled brackets of the non-terminal nodes
of $T$.\\
$Cross(U, V)$ denotes the subset of brackets from $U$ that cross at least one
bracket in $V$.\\
$O_i$ represents a tree in the learned corpus.\\
$T_i$ is a tree in $TEST$, the original corpus. \cite{bib:led}
\end{tabular}
\end{center}
\caption{Formulas of the different metrics}
\label{fig:metrics}
\end{figure*}

The results of the ABL:incr system depend on the order of the sentences in the
corpus and the two probabilistic ABL systems sometimes choose constituents at
random (i.e. when more combinations of constituents have the same probability).
Therefore, the ABL:incr system is applied to ten differently ordered versions
of the corpora. Likewise, we applied the probabilistic ABL systems ten times to
the corpora to account for the random selection of constituents. The mean
results are shown in table~\ref{tab:results} and the standard deviations are
shown in brackets.

Three different metrics are used to compare the results of the different
algorithms:
\begin{description}
\item [NCBP] Non-Crossing Brackets Precision: the percentage of learned
constituents that do not overlap any constituents in the original corpus.
\item [NCBR] Non-Crossing Brackets Recall: the percentage of original
constituents that do not overlap any constituents in the learned corpus.
\item [ZCS] Zero-Crossing Sentences: the percentage of sentences that do not
have any overlapping constituents.
\end{description}
The formulas describing the metrics can be found in figure~\ref{fig:metrics}.

\subsection{Test Results}

\begin{table*}
\begin{center}

\begin{tabular}{|l||r@{\hspace{3pt}}r|r@{\hspace{3pt}}r|r@{\hspace{3pt}}r||
r@{\hspace{3pt}}r|r@{\hspace{3pt}}r|r@{\hspace{3pt}}r|}\hline
\rule[0in]{0in}{0.20in}\ignorespaces
\rule[-0.10in]{0in}{0.10in}\ignorespaces
                   & \multicolumn{6}{c||}{Results ATIS corpus} &
\multicolumn{6}{c|}{Results OVIS corpus} \\\hline
\rule[0in]{0in}{0.20in}\ignorespaces
\rule[-0.10in]{0in}{0.10in}\ignorespaces
&\multicolumn{2}{c|}{NCBP}&\multicolumn{2}{c|}{NCBR}&\multicolumn{2}{c||}{ZCS}&
 \multicolumn{2}{c|}{NCBP}&\multicolumn{2}{c|}{NCBR}&\multicolumn{2}{c|}{ZCS}\\
\hline
left       & 32.60 &        & 76.82 &        & 1.12  &        &
             51.23 &        & 73.17 &        & 25.22 &        \\
right      & 82.70 &        & 92.91 &        & 38.83 &        &
             75.85 &        & 86.66 &        & 48.08 &        \\

ABL:incr   & 82.55 & (0.80) & 82.98 & (0.78) & 17.15 & (1.17) &
             88.69 & (1.11) & 83.90 & (1.61) & 45.13 & (4.12) \\
ABL:leaf   & 82.20 & (0.30) & 82.65 & (0.29) & 21.05 & (0.76) &
             85.70 & (0.01) & 79.96 & (0.02) & 30.87 & (0.07) \\
\rule[-0.10in]{0in}{0.10in}\ignorespaces
ABL:branch & 86.04 & (0.10) & 87.11 & (0.09) & 29.01 & (0.00) &
             89.39 & (0.00) & 84.90 & (0.00) & 42.05 & (0.02) \\\hline
\end{tabular}
\end{center}
\caption{Results ATIS and OVIS corpora}
\label{tab:results}
\end{table*}

The results of applying the ABL algorithms and the baseline systems on the ATIS
and OVIS corpora can be found in table~\ref{tab:results}. The ATIS corpus is
predominantly right branching as can be seen in the results, while the OVIS
corpus is right branching to a lesser degree.

The ABL:branch method performs significantly better on all metrics on both the
ATIS and OVIS corpora compared to the other ABL methods. The only deviating ABL
result is the ZCS of the ABL:incr system in the OVIS corpus.

It is interesting to see that the ABL:incr system outperforms ABL:leaf. In the
ABL:incr system, incorrectly learned constituents will never be corrected.
The idea behind the probabilistic methods (including ABL:leaf) is that
incorrect constituents will be corrected because they will receive lower
probabilities. Apparently, the statistics used in the ABL:leaf method do not
provide enough information to make a correct bracket selection, whereas
ABL:branch does.

The standard deviation of the results of the ABL:incr system is quite large,
which indicates that the order of the sentences in the corpus is important. On
the other hand, selecting random constituents (when multiple constituents have
the same probability) in the probabilistic ABL systems yields less varying
results. When a more specific probabilistic method is used (ABL:branch), the
variance in results is almost zero.

The right branching system outperforms ABL systems on several metrics (the
ABL:branch method almost keeps up however). This could be expected, since the
ATIS is predominantly right branching (the OVIS corpus to a lesser degree).
Therefore, the right branching system should perform well. The ABL systems do
not have a directional branching method built-in; furthermore, the ABL systems
have no adjustable parameters that can be used to tune the algorithms to the
specific corpora. This means that the right branching system does not perform
well on a corpus of a predominantly left branching language (for example
Japanese), while we expect that ABL does.

It is difficult to compare the results of the ABL model against other methods,
since often different corpora or metrics are used. The methods described by
\newcite{bib:iorfpbc} comes reasonably close to ours. The \emph{unsupervised}
method learns structure on plain sentences from the ATIS corpus resulting in
37.35~\% non-crossing brackets precision, while the \emph{unsupervised} ABL
significantly outperforms this method, reaching 86.04~\% precision. Only their
\emph{supervised} version results in a slightly higher precision of 90.36~\%.

\section{Future Work}

Although the overall result of ABL algorithm is slightly disappointing, some of
the results of the different ABL algorithms are encouraging. We expect future
extensions of the ABL system to improve performance.

This section contains some future extensions to the ABL implementations
described in this paper. These extensions solve certain problems the current ABL
systems have. First, we take a look at different alignment schemes, that may be
used as an alternative to the string edit distance algorithm. Then an
alternative probabilistic model for bracket selection is considered. Finally,
we discuss two methods that generate more possible constituents by adding
context or weakening exact match.

\subsection{Different Alignment Schemes}

The edit distance algorithm that is used to find identical parts in sentences
sometimes finds alignments that generate unintended constituents. If we consider
the sentences~1 and 2 in figure~\ref{fig:unintended}, the edit distance aligns
these sentences as in sentences~3 and 4. Unfortunately, the alignment in
sentences~3 and 4 generate unintended constituents. The most preferred
alignment is shown in sentences~7 and 8, since \textit{San Francisco} and
\textit{Dallas} are grouped and receive the same type.

\begin{figure}[ht]
\begin{center}
\begin{tabular}{l}
\textbf{1} \textit{from San Francisco to Dallas}\\
\textbf{2} \textit{from Dallas to San Francisco}\\[.05cm]
\hline
\textbf{3} \textit{from ()$_{X_1}$ San Francisco (to Dallas)$_{X_2}$}\\
\textbf{4} \textit{from (Dallas to)$_{X_1}$ San Francisco ()$_{X_2}$}\\[.05cm]
\hline
\textbf{5} \textit{from (San Francisco to)$_{X_3}$ Dallas ()$_{X_4}$}\\
\textbf{6} \textit{from ()$_{X_3}$ Dallas (to San Francisco)$_{X_4}$}\\[.05cm]
\hline
\textbf{7} \textit{from (San Francisco)$_{X_5}$ to (Dallas)$_{X_6}$}\\
\textbf{8} \textit{from (Dallas)$_{X_5}$ to (San Francisco)$_{X_6}$}
\end{tabular}
\end{center}
\vspace{-.3cm}
\caption{Unintended constituents}
\label{fig:unintended}
\end{figure}

This problem occurs every time the algorithm aligns words that are ``too far
apart''. The relative distance between the two \textit{San Francisco}s in the
two sentences is much larger than the relative distance between the word
\textit{to} in both sentences.

There are two possible solutions to this problem. The first solution is to
change the cost function of the edit distance algorithm to let it select the
better alignment. Another solution is to simply generate all possible
alignments (using a completely different alignment algorithm) and let the
bracket selection phase of the algorithm select the best alignment.

A better cost function should be biased towards alignments with a small relative
distance. This can be accomplished by letting the cost function return a high
cost when the difference of the relative offsets of the words is large. The
relative distance between the two \textit{San Francisco}s in sentences~3 and 4
in figure~\ref{fig:unintended} is larger compared to the relative distance
between the two \textit{to}s in sentences~7 and 8. Therefore the total edit cost
of sentences~7 and 8 will be less than the edit cost of sentences~3 and 4 or
sentences~5 and 6.

When all possible alignments are generated, there is a large probability the
intended alignment will also be found. Unfortunately, it is not known which of
the possible alignments is the intended. The bracket selection phase should
select the best constituents from all possible constituents that are generated
by the alignment learning phase. When the probabilistic bracket selection
methods are used, we just have to assume that the intended alignment generates
more probable constituents, or equivalently that the intended alignment
contains constituents that occur more often in the corpus than the unintended
constituents.

\subsection{Alternative Probabilistic Models}

The results of the application of the algorithm on the ATIS corpus show that
the probabilistic methods generate less fluctuating results (i.e. they have a
smaller standard deviation) than the non-probabilistic method. The results of
the ABL:branch method show less deviation than the results of the ABL:leaf
method. Furthermore, the ABL:branch method generates the best results. These
results indicate that a more specific probabilistic method works better in
bracket selection.

As future research, a DOP-like probabilistic method \cite{bib:bgaebtol} of
bracket selection will be implemented. In this approach, all possible
constituents are broken into fragments (cf. elementary subtrees in DOP). The
probability of such a fragment $f$ is:
\[
P(f|root(f)=r)=\frac{|f|}{\sum_{f':root(f')=r}|f'|}
\]
When combining these fragments, the structure of constituent $c$ can be
generated. This can be seen as a derivation of the structure of the
constituent. The probability of a derivation $f_1\circ \ldots\circ f_n$
is:
\[
P(f_1\circ \ldots\circ f_n)=\prod_i P(f_i)
\]
Usually, there is more than one way of deriving the structure of the
constituent. The probability of the constituent is now the combination of all
derivations that yield the structure of $c$:
\[
P(c)=\sum_{d\mbox{\scriptsize{ derives }}c}P(d)
\]

The probabilistic model of DOP does not only use the counts of terminals or
non-terminals like in the ABL:leaf and ABL:branch methods, but also uses the
internal structure of constituents. This yields a much more specific
stochastic model.

The ABL:branch bracket selection method is comparable to the probabilistic
model of SCFGs (cf. \cite{bib:profl}), where probabilities of context-free
grammar rules (terminals and their root non-terminal) are used. Since DOP
clearly outperforms SCFGs (as shown in \cite{bib:elwspmonl}), a DOP model in
bracket selection is expected to increase performance.

\subsection{Adding Context}
\label{s:ac}

Some problematic cases exist where ABL might seem unable to learn the correct
syntactic type. Consider sentences as in figure~\ref{fig:wellmeat}. The ABL
algorithm finds that \textit{morning} and \textit{nonstop} are of the same
type, since the rest of the two sentences is identical. However,
\textit{morning} is tagged as \textit{NN} (a noun) and \textit{nonstop} as
\textit{JJ} (an adjective).

\begin{figure}[ht]
\begin{center}
\begin{tabular}{l}
\textit{Show me the (morning)$_{X_1}$ flights}\\
\textit{Show me the (nonstop)$_{X_1}$ flights}
\end{tabular}
\end{center}
\vspace{-.3cm}
\caption{Inconsistent syntactic types}
\label{fig:wellmeat}
\end{figure}

On the other hand, one might argue these words \emph{are} of the same type,
exactly because they occur in the same context. Both words might be seen as
some sort of adjective phrase.

This is a difference between syntactic type and functional type.
\textit{Morning} and \textit{nonstop} have a different syntactic type, a noun
and an adjective respectively, but both modify the noun \textit{flights}, i.e.
they have the same functional type. ABL finds the functional type, while the
words are tagged according to their syntactic type.

Since the overall use of the two words differs greatly, they occur in different
contexts. \textit{Morning} may in general occur in places where nouns or noun
phrases belong, while \textit{nonstop} may not. This discrepancy can be used to
differentiate between the two words. Instead of giving the words only one type,
they get two: one type describes the context (i.e. the functional type), which
is the same for both words and the other type describes the syntactic type,
which is different for both words.

The distribution of syntactic types combined with functional types can be used
to find words that belong to the same syntactic type. \textit{Morning} and
\textit{nonstop} have the same functional type but different syntactic types,
since \textit{morning} normally occurs in other contexts than \textit{nonstop}.

Since the functional type should describe the context of a constituent, merging
of constituents as in figure~\ref{fig:learn1} and figure~\ref{fig:learn2}
should only apply to functional types. Merging of syntactic types is only done
when the distribution of the constituents is similar enough. This effectively
loosens the assumption that constituents in a certain context have the same
(syntactic \emph{and} functional) type.

\subsection{Weakening Exact Match}

The algorithm described in this paper is unable to learn any structure when two
completely different sentences are compared. (This is not an insurmountable
problem, since other sentences can be used to learn structure on the two
sentences.) 

The algorithms described so far all try to align using exactly matching words.
Sometimes this is a too strong demand; it is enough to match similar words.
Imagine sentences~1 and 2 in figure~\ref{fig:equivalent}, which are completely
distinct. The standard ABL learning methods would conclude both are sentences,
but no more structure will be found. When the algorithm knows \textit{Book} and
\textit{List} are words of the same type (representing verbs), it would find
the structures in sentences~3 and 4 where the type $X_1$ represents a noun
phrase.

\begin{figure}[ht]
\begin{center}
\begin{tabular}{l}
\textbf{1} \textit{Book Delta flight 430}\\
\textbf{2} \textit{List the cost for limousines}\\[.05cm]
\hline
\textbf{3} \textit{Book (Delta flight 430)$_{X_1}$}\\
\textbf{4} \textit{List (the cost for limousines)$_{X_1}$}
\end{tabular}
\end{center}
\vspace{-.3cm}
\caption{Sentences without identical words}
\label{fig:equivalent}
\end{figure}

An obvious way of implementing this is by using \emph{equivalence classes}. (See
for example \cite{bib:diapcfasc}.) Words that are closely related are grouped
together in the same class. Words in the same equivalence class are similar
enough to be aligned.

A big advantage of equivalence classes is that they can be learned in an
unsupervised way. Even when the algorithm is extended with equivalence classes,
it still does not need to be initialised with structured training data.

\section{Conclusion}

In this paper a new language learning algorithm based on aligning sentences is
introduced. It uses distinctions between sentences to find possible
constituents during the alignment learning phase and selects the most probable
constituents afterwards in the bracket selection phase.

Three instances of the algorithm have been applied to the ATIS corpus (716
sentences) and the OVIS corpus (6,797 sentences). The alignment learning phase
in the three systems is the same, but the bracket selection phase differs.
ABL:incr assumes earlier constituents learned are correct, ABL:leaf and
ABL:branch select brackets based on their probability. The ABL:branch method,
which uses the most specific stochastic model, performs best.

The ABL algorithm is an unsupervised grammar learning and bootstrapping system.
It uses plain sentences to learn structure, so neither pre-tagged, pre-labelled
nor pre-bracketed sentences are used. There is no need to train the system on a
structured corpus and the system does not use any parameters that need to be
set.

Possible constituents are found by looking at an unlimited context. The
algorithm does not use a fixed window size. Alignments (and thus constituents)
of arbitrarily large size may be considered.

Furthermore, the ABL algorithm is able to learn recursion from a finite set of
sentences.

\bibliographystyle{acl}
\bibliography{paper}

\begin{thebibliography}{}

\bibitem[\protect\citename{Baker}1979]{bib:tgfsr}
J.~K. Baker.
\newblock 1979.
\newblock Trainable grammars for speech recognition.
\newblock In J.~J. Wolf and D.~H. Klatt, editors, {\em Speech Communication
  Papers for the Ninety-seventh Meeting of the Acoustical Society of America},
  pages 547--550.

\bibitem[\protect\citename{Bod}1995]{bib:elwspmonl}
Rens Bod.
\newblock 1995.
\newblock {\em Enriching Linguistics with Statistics: Performance Models of
  Natural Language}.
\newblock {Ph.D.} thesis, Universiteit van Amsterdam.

\bibitem[\protect\citename{Bod}1998]{bib:bgaebtol}
Rens Bod.
\newblock 1998.
\newblock {\em Beyond Grammar --- An Experience-Based Theory of Language}.
\newblock Stanford, CA: {CSLI} Publications.

\bibitem[\protect\citename{Bonnema \bgroup et al.\egroup }1997]{bib:admfsi}
R.~Bonnema, R.~Bod, and R.~Scha.
\newblock 1997.
\newblock A {DOP} model for semantic interpretation.
\newblock In {\em Proceedings of the {A}ssociation for {C}omputational
  {L}inguistics/{E}uropean {C}hapter of the {A}ssociation for {C}omputational
  {L}inguistics, Madrid}, pages 159--167. Sommerset, NJ: Association for
  Computational Linguistics.

\bibitem[\protect\citename{Booth}1969]{bib:profl}
T.~Booth.
\newblock 1969.
\newblock Probabilistic representation of formal languages.
\newblock In {\em Conference Record of 1969 Tenth Annual Symposium on Switching
  and Automata Theory}, pages 74--81.

\bibitem[\protect\citename{Brill}1993]{bib:agiapftatba}
Eric Brill.
\newblock 1993.
\newblock Automatic grammar induction and parsing free text: A
  transformation-based approach.
\newblock In {\em Proceedings of the {A}ssociation for {C}omputational
  {L}inguistics}, pages 259--265.

\bibitem[\protect\citename{Caraballo and Charniak}1998]{bib:nfomfbfpcp}
Sharon~A. Caraballo and Eugene Charniak.
\newblock 1998.
\newblock New figures of merit for best-first probabilistic chart parsing.
\newblock {\em Computational Linguistics}, 24(2):275--298.

\bibitem[\protect\citename{Chen}1995]{bib:bgiflm}
Stanley~F. Chen.
\newblock 1995.
\newblock Bayesian grammar induction for language modeling.
\newblock In {\em Proceedings of the {A}ssociation for {C}omputational
  {L}inguistics}, pages 228--235.

\bibitem[\protect\citename{Cook \bgroup et al.\egroup }1976]{bib:gibhc}
C.~M. Cook, A.~Rosenfeld, and A.~R. Aronson.
\newblock 1976.
\newblock Grammatical inference by hill climbing.
\newblock {\em Informational Sciences}, 10:59--80.

\bibitem[\protect\citename{Daelemans}1995]{bib:mblaap}
Walter Daelemans.
\newblock 1995.
\newblock Memory-based lexical acquisition and processing.
\newblock In P.~Steffens, editor, {\em Machine Translation and the Lexicon},
  volume 898 of {\em Lecture Notes in Artificial Intelligence}, pages 85--98.
  Berlin: Springer Verlag.

\bibitem[\protect\citename{de Marcken}1996]{bib:ula}
Carl~G. de~Marcken.
\newblock 1996.
\newblock {\em Unsupervised Language Acquisition}.
\newblock {Ph.D.} thesis, Department of Electrical Engineering and Computer
  Science, Massachusetts Institute of Technology, Cambridge, MA, sep.

\bibitem[\protect\citename{Finch and Chater}1992]{bib:bscusm}
Steven Finch and Nick Chater.
\newblock 1992.
\newblock Bootstrapping syntactic categories using statistical methods.
\newblock In Walter Daelemans and David Powers, editors, {\em Background and
  Experiments in Machine Learning of Natural Language: Proceedings First {SHOE}
  Workshop}, pages 229--235. Institute for Language Technology and {AI},
  Tilburg University, The Netherlands.

\bibitem[\protect\citename{Gr{\"u}nwald}1994]{bib:giatmdlp}
Peter Gr{\"u}nwald.
\newblock 1994.
\newblock A minimum description length approach to grammar inference.
\newblock In G.~Scheler, S.~Wernter, and E.~Riloff, editors, {\em
  Connectionist, Statistical and Symbolic Approaches to Learning for Natural
  Language}, volume 1004 of {\em Lecture Notes in {AI}}, pages 203--216.
  Berlin: Springer Verlag.

\bibitem[\protect\citename{Harris}1951]{bib:misl}
Zellig Harris.
\newblock 1951.
\newblock {\em Methods in Structural Linguistics}.
\newblock Chicago, IL: University of Chicago Press.

\bibitem[\protect\citename{Kehler and Stolcke}1999]{bib:ulinlp}
Andrew Kehler and Andreas Stolcke.
\newblock 1999.
\newblock Preface.
\newblock In A.~Kehler and A.~Stolcke, editors, {\em Unsupervised Learning in
  Natural Language Processing}. Association for Computational Linguistics.
\newblock Proceedings of the workshop.

\bibitem[\protect\citename{Lari and Young}1990]{bib:teoscfgutioa}
K.~Lari and S.~J. Young.
\newblock 1990.
\newblock The estimation of stochastic context-free grammars using the
  inside-outside algorithm.
\newblock {\em Computer Speech and Language}, 4:35--56.

\bibitem[\protect\citename{Magerman and Marcus}1990]{bib:pnlumis}
D.~Magerman and M.~Marcus.
\newblock 1990.
\newblock Parsing natural language using mutual information statistics.
\newblock In {\em Proceedings of the National Conference on Artificial
  Intelligence}, pages 984--989. Cambridge, MA: MIT Press.

\bibitem[\protect\citename{Marcus \bgroup et al.\egroup }1993]{bib:balacoetpt}
M.~Marcus, B.~Santorini, and M.~Marcinkiewicz.
\newblock 1993.
\newblock Building a large annotated corpus of english: the {P}enn treebank.
\newblock {\em Computational Linguistics}, 19(2):313--330.

\bibitem[\protect\citename{Pereira and Schabes}1992]{bib:iorfpbc}
F.~Pereira and Y.~Schabes.
\newblock 1992.
\newblock Inside-outside reestimation from partially bracketed corpora.
\newblock In {\em Proceedings of the {A}ssociation for {C}omputational
  {L}inguistics}, pages 128--135, Newark, Delaware.

\bibitem[\protect\citename{Powers}1997]{bib:mlofnl}
David M.~P. Powers.
\newblock 1997.
\newblock Machine learning of natural language.
\newblock {A}ssociation for {C}omputational {L}inguistics/{E}uropean {C}hapter
  of the {A}ssociation for {C}omputational {L}inguistics Tutorial Notes,
  Madrid, Spain.

\bibitem[\protect\citename{Redington \bgroup et al.\egroup
  }1998]{bib:diapcfasc}
Martin Redington, Nick Chater, and Steven Finch.
\newblock 1998.
\newblock Distributional information: A powerful cue for acquiring syntactic
  categories.
\newblock {\em Cognitive Science}, 22(4):425--469.

\bibitem[\protect\citename{Sima'an}1999]{bib:led}
Khalil Sima'an.
\newblock 1999.
\newblock {\em Learning Efficient Disambiguation}.
\newblock {Ph.D.} thesis, Institute for Language, Logic and Computation,
  Universiteit Utrecht.

\bibitem[\protect\citename{Stolcke and Omohundro}1994]{bib:ipgbbmm}
Andreas Stolcke and Stephen Omohundro.
\newblock 1994.
\newblock Inducing probabilistic grammars by bayesain model merging.
\newblock In {\em Second International Conference on Grammar Inference and
  Applications}, pages 106--118. Berlin: Springer Verlag.
\newblock Alicante, Spain.

\bibitem[\protect\citename{Viterbi}1967]{bib:ebfccaaaoda}
A.~Viterbi.
\newblock 1967.
\newblock Error bounds for convolutional codes and an asymptotically optimum
  decoding algorithm.
\newblock {\em Institute of Electrical and Electronics Engineers Transactions
  on Information Theory}, 13:260--269.

\bibitem[\protect\citename{Wagner and Fischer}1974]{bib:tstscp}
Robert~A. Wagner and Michael~J. Fischer.
\newblock 1974.
\newblock The string-to-string correction problem.
\newblock {\em Journal of the Association for Computing Machinery},
  21(1):168--173, jan.

\bibitem[\protect\citename{Wolff}1982]{bib:ladcag}
J.~G. Wolff.
\newblock 1982.
\newblock Language acquisition, data compression and generalization.
\newblock {\em Language \& Communication}, 2:57--89.

\end{thebibliography}

\end{document}